# Reinforcement Learning for Self-Healing Material Systems

Maitreyi Chatterjee[1*], Devansh Agarwal[1*], Biplab Chatterjee[3]
[1]Cornell University, Ithaca, NY
[3]Ground Handling, AI Airport Services Ltd, Kolkata, India
* Equal Contribution

**ABSTRACT**

The transition to autonomous material systems necessitates adaptive control methodologies to maximize structural longevity. This study frames the self-healing process as a Reinforcement Learning (RL) problem within a Markov Decision Process (MDP), enabling agents to autonomously derive optimal policies that efficiently balance structural integrity maintenance against finite resource consumption. A comparative evaluation of discrete-action (Q-learning, DQN) and continuous-action (TD3) agents in a stochastic simulation environment revealed that RL controllers significantly outperform heuristic baselines, achieving near-complete material recovery. Crucially, the TD3 agent utilizing continuous dosage control demonstrated superior convergence speed and stability, underscoring the necessity of fine-grained, proportional actuation in dynamic self-healing applications.

**NOMENCLATURE**
**RL,** Reinforcement Learning
**Q-learning,** Quality-learning
**DQN,** Deep Q-Network

## 1. INTRODUCTION

The capacity for autonomous repair marks a paradigm shift in materials science, enabling extended service lifetimes and reduced reliance on external maintenance [1, 2]. While traditional self-healing methods, like microcapsule release, are limited to pre-programmed responses, recent advances in Machine Learning [3, 4, 5, 6] allow agents to learn adaptive repair strategies directly from experience [7, 8]. We model self-healing as a **Markov Decision Process (MDP)**, where the agent balances durability against resource use in stochastic and continuous environments. Our contributions are:

- A **Reinforcement learning control framework** tailored for self-healing systems.
- A **Stochastic simulation environment** capturing essential damage–repair dynamics.
- **Comparative evaluation of Q-learning, DQN, and TD3**, demonstrating superior recovery and efficiency over heuristic and random baselines.

## 2. COMPUTATIONAL METHODS: REINFORCEMENT LEARNING FOR SELF-HEALING

The goal is for the RL agent to learn a policy that maximizes the effectiveness and longevity of the self-healing process.

### 2.1. System Architecture

Inspired by prior work in sensor-actuator systems [8], our architecture includes:

- **Self-healing material** with embedded healing agents, sensors, and micro-actuators.
- **Sensor arrays** that monitor damage initiation and progression.
- **RL agent** that processes sensor input, selects actions, and updates its policy.
- **Actuators** that apply localized interventions (e.g., heating, chemical release, compression).

### 2.2. Reinforcement Learning Formulation

The self-healing problem can be framed as a Markov Decision Process (MDP)

- **State (S):** captures damage severity, material properties, healing agent status, and environment.
- **Actions (A):** apply chemical release, thermal activation, mechanical compression, or no action.
- **Reward (R):** balances integrity recovery against healing cost and resource use.

- **Policy (π):** maps states to actions to maximize long-term integrity.

### 2.3. Experimental Setup

We simulated a self-healing environment inspired by vascular and encapsulated systems [7]. Damage accumulated randomly over time and could be reduced by healing actions. At each timestep, the agent observed current integrity, selected an action, and received a reward balancing recovery, cost, and penalties. Simulations began with integrity = **0.91,** ran for **120** steps, and results were averaged across **10 runs**

#### 2.3.1 Discrete-Action Agents.

We implemented tabular Q-learning and a neural-network-based Deep Q-Network (DQN). Both agents operated over a discrete action space with three options - **Chemical release** (high cost, high healing), **Thermal activation** (moderate cost, moderate healing), or **Do nothing** (zero cost, penalized by damage severity).

The Q-Learning agent uses an **ε-greedy** policy with the standard Bellman equation:

$$Q(s,a) = Q(s,a) + \alpha \left[ r + \gamma \max_{a'} Q(s',a') - Q(s,a) \right]$$

The Deep Q-Network (DQN) has the following stability enhancements:

1. **Experience replay buffer** (size ≈ 10,000), from which minibatches were sampled uniformly for training.
2. **Target network updated periodically** to stabilize bootstrapped value estimates.

#### 2.3.2 Continuous-Action Agent.

To better approximate real-world scenarios where healing interventions may be applied at varying intensities, we extended the environment to a continuous action space a∈[0,1], representing the dosage of chemical release. Healing and cost scaled proportionally with the action magnitude. For this setting, we trained **Twin Delayed Deep Deterministic Policy Gradient (TD3)** agent with twin critics, target networks, and policy smoothing noise. TD3 enables stable learning in continuous domains and avoids overestimation bias common in actor–critic methods. Deploying reinforcement learning (RL) in self-healing materials is challenged by limited interaction budgets, since real experiments are costly and slow. To address this, the study evaluated data-efficient strategies like **Prioritized Experience Replay (PER)** and **Transfer Learning**

## 3. RESULTS AND DISCUSSION

The Q-Learning agent develops a context-sensitive policy, learning to save resources when integrity is stable and respond only when necessary (**Fig. 1**).

- **Chemical release** was reserved for severe damage.
- **Thermal activation** handled moderate damage.
- **No action** was used when healing was unnecessary (most frequent).

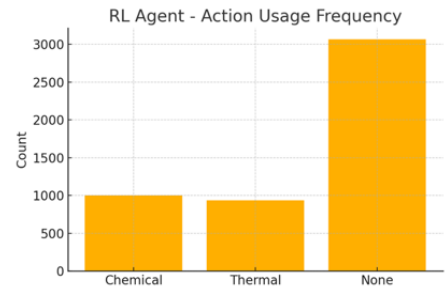

**Fig. 1: Action Usage Frequency**

This behavior aligns with resource-constrained planning observed in biological systems and energy-aware control agents [7]

### 3.1. Final Simulation Results

Starting from integrity 0.91, Q-learning and DQN restored health above 0.98 within 20–30 steps and maintained ≈0.99 for 120 steps. TD3 recovered even faster, reaching ≈1.0 within 5–6 steps and holding steady thereafter **(Fig. 2)**. This reflects a resource-aware strategy: minor damage is tolerated, while decisive healing is applied only when necessary.

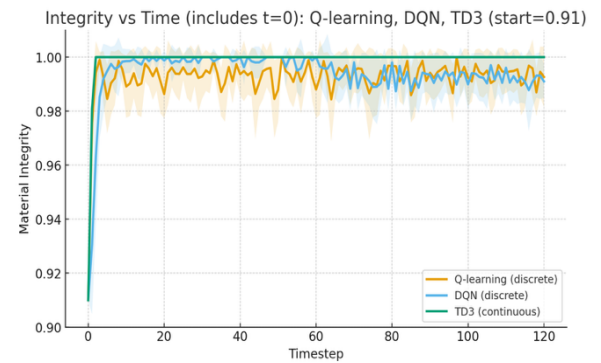

**Fig. 2: Material Integrity trajectory**

### 3.2. Performance comparison with baselines

We compare the RL agent to a random agent and a heuristic agent (heals when integrity < 0.8) as well as a **proportional–integral (PI)-style** adaptive controller. The random agent fails to recover material health effectively, while the heuristic overuses healing actions, leading to reduced resource efficiency. The adaptive controller recovered integrity to ≈0.96 on average. In contrast, the RL agent achieves near-complete recovery with lower supply usage.

**Table 1: Performance comparison with baseline**

| Agent | Avg reward | Avg Final Integrity | Supply Used |
|---|---|---|---|
| Qlearnig | -22.57 | 0.997 ± 0.006 | 58 |
| Heuristic | -25.12 | 0.860 ± 0.020 | High |
| Random | 27.98 | 0.710 ± 0.030 | High |
| DQN | -23.10 | 0.992 ± 0.011 | ~60 |
| TD3 | -21.85 | 0.999 ± 0.004 | ~55 |
| Adaptive | -24.90 | 0.962 ± 0.014 | High |

Over 120 timesteps × 10 runs, Q-learning consistently achieved high stability, with final integrity ≈ **0.997 ± 0.006.** DQN, despite incorporating experience replay and target networks, showed greater variability (≈ 0.992 ± 0.011), with occasional dips in recovery. The contrast illustrates a trade-off:

1. Q-learning is optimal for compact, tabular environments where state–action spaces are tractable.
2. DQN offers scalability to richer sensor states (e.g., continuous FEM models, real actuator signals), where tabular methods are infeasible despite their stability.

### 3.3. Extending to Continuous Control with TD3

To better approximate real-world healing systems where dosage is adjustable, we extended to continuous action space (dosage ∈ [0,1]). Using Twin Delayed Deep Deterministic Policy Gradient (TD3), the agent rapidly restored integrity to ≈ 1.0 within the first 5–6 steps and maintained it with minimal variance across runs. This demonstrates the advantage of fine-grained dosing control, avoiding oscillations from over- or under-correction seen in discrete policies.

We extended the self-healing environment to account for probabilistic outcomes. Each healing action may only succeed with some probability and, if successful, may restore only a fraction of the intended effect. Formally, the realized healing combines a Bernoulli success indicator and a Beta-distributed partial efficacy factor. In experiments, RL agents remained effective under uncertainty, maintaining near-complete integrity while adopting more conservative, risk-aware strategies. For example, TD3 achieved **0.993±0.006** integrity across **30 runs**, with the probability of failure reduced from **8.1%** (deterministic model) to **2.4%**.

A lightweight grid-based surrogate model was used to mimic FEM-like stress and damage dynamics. Stress was estimated with a Laplacian operator, causing damage growth above a threshold and reduction under localized healing. Starting from integrity 0.91, the greedy controller maintained ~0.99 over 120 steps, while "no control" drifted downward. An oracle using hidden damage set an upper bound near 1.0, with heatmaps (Fig. 3) confirming targeted reduction of central defects.

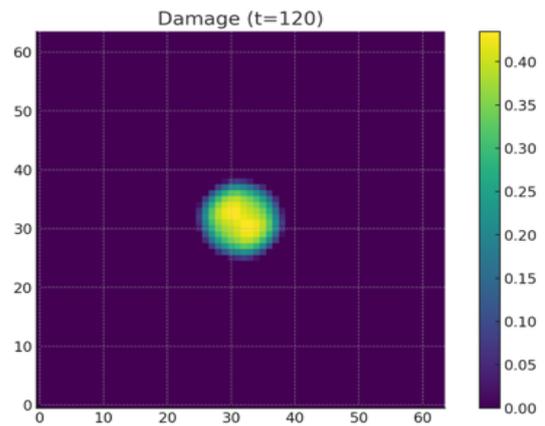

**Fig. 3: Damage Heatmap at t=120**

### 3.4 Data Efficiency through Prioritized Replay

Under a constrained budget of 60 interaction steps (10 runs, initial integrity = 0.91):

- **Baseline DQN** with uniform replay recovers slowly, reaching only ~0.95 integrity by step 60.
- **DQN + Prioritized Experience Replay** (PER) improves efficiency, achieving ~0.98 integrity by prioritizing informative transitions.
- **DQN + Transfer Learning** benefits from offline prefill, showing better early stability, though it may plateau at slightly lower integrity levels.

### 3.5 Insights and Broader Impacts

A key challenge in real-world RL-driven healing is avoiding premature or excessive actuation. This can be mitigated with multi-layered safety interlocks: hardware cutoffs (fuses, dose caps), firmware watchdogs (cooldowns, budget limits), and reward penalties at the policy level.

From a computational perspective, RL inference is lightweight: tabular policies are trivial to run, and compact neural networks (1–2 hidden layers, 32–64 units) can execute within milliseconds on embedded microcontrollers. Unlike bio-inspired self-healing mechanisms (capsule release, vascular networks, or shape-memory polymers), which are rigid and pre-programmed, RL enables adaptive, context-aware control—learning when, where, and how much to act under uncertain damage conditions.

The supply–reward analysis **(Fig. 4)** shows that the agent learns a balanced cost–benefit strategy, allocating healing resources efficiently while avoiding excessive use. This demonstrates that reinforcement learning can achieve resource-conscious control, ensuring material stability without overconsumption.

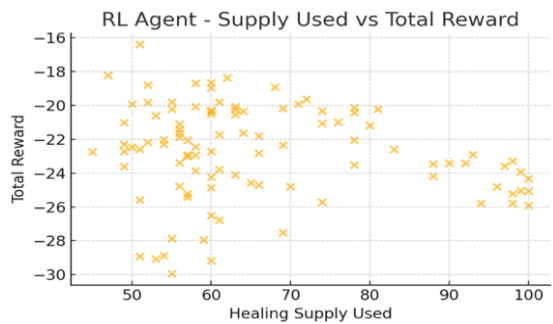

**Fig. 4: Supply vs Reward Scatter Plot**

## 4. LIMITATIONS AND FUTURE WORK

The healing environment was implemented using a surrogate stress–damage grid rather than full finite element simulations. While this allows efficient training, it may not capture all nonlinearities and multi-physics couplings present in real materials. Future work aims to transition from the lightweight FEM-lite surrogate to full finite-element solvers (e.g., COMSOL, Abaqus, FEniCSx) for physically accurate stress and crack modeling. Combining such high-fidelity simulations with surrogate-based pretraining and prioritized replay can maintain sample efficiency while improving realism.

A practical next step is the translation of RL-driven control policies into laboratory prototypes. Self-healing composites can be instrumented with sensor–actuator networks, where embedded strain gauges, fiber Bragg gratings, or capacitive sensors detect local stress concentrations and microcrack initiation. These signals can be processed by low-power microcontrollers (e.g., STM32, Arduino-class, or Raspberry Pi Pico), which host lightweight RL inference models.